\documentclass{article}
\usepackage{spconf,amsmath,graphicx}

\usepackage{cite}
\usepackage{amsmath,amssymb,amsfonts}
\usepackage{algorithmic}
\usepackage{graphicx}
\usepackage{textcomp}
\def\BibTeX{{\rm B\kern-.05em{\sc i\kern-.025em b}\kern-.08em
    T\kern-.1667em\lower.7ex\hbox{E}\kern-.125emX}}

\usepackage{subcaption}
\usepackage{setspace}
\usepackage{type1cm}
\usepackage{caption}
\usepackage{bm}

\usepackage{array}

\usepackage{fixltx2e}

\usepackage{nicefrac}       % compact symbols for 1/2, etc.
\usepackage{array}
\usepackage{makecell}

\usepackage{hyperref}

\usepackage{paralist}

\usepackage{tikz}
\usetikzlibrary{matrix,positioning,shapes,shadows,arrows,calc,decorations.pathreplacing}
\usetikzlibrary{arrows.meta}

\tikzstyle{blocky} = [draw, line width=1.2pt,fill=white, rectangle, 
minimum height=2.5em, minimum width=4.5em, rounded corners]    

\tikzstyle{blocky2} = [draw, line width=1.2pt,fill=red!10, rectangle, 
minimum height=2.5em, minimum width=18.5em, rounded corners]

\definecolor{darkred1}{RGB}{228,26,28}
\definecolor{darkblue1}{RGB}{55,126,184}
\definecolor{darkgreen1}{RGB}{77,175,74}
\title{Diffusion Maps meet Nystr\"om}
\name{N. Benjamin Erichson$^{\star}$ \qquad Lionel Mathelin$^{\dagger}$ \qquad Steven Brunton$^{\star}$ \qquad Nathan Kutz$^{\star}$}

\address{$^{\star}$ Applied Mathematics, University of Washington, Seattle, USA \\
	$^{\dagger}$ LIMSI-CNRS (UPR 3251), Campus Universitaire d'Orsay, 91405 Orsay cedex, France}

\begin{document}
%\ninept
%
\maketitle
\begin{abstract}
Diffusion maps are an emerging data-driven technique for non-linear dimensionality reduction, which are especially useful for the analysis of coherent structures and nonlinear embeddings of dynamical systems.
However, the computational complexity of the diffusion maps algorithm scales with the number of observations. Thus, long time-series data presents a significant challenge for fast and efficient embedding. 
We propose integrating the Nystr\"om method with diffusion maps in order to ease the computational demand.
We achieve a speedup of roughly two to four times when approximating the dominant diffusion map components. 
%and achieve a speedup of roughly two to four times.
%We demonstrate the algorithm on synthetic data and the chaotic Lorenz system, which is among the simplest and well-known chaotic dynamical system.
%

%by taking advantage of the low-rank structure and symmetry of kernel functions.
%

%
%We demonstrate the algorithm on synthetic data and the chaotic Lorenz system, which is among the simplest and well-known chaotic dynamical system.
%
%Further, we enable the fast exploration of the hyper-parameter space to tune diffusion map.   

%- summary:  Diffu maps are a emerging dim. reduct which is limited by being unscalable.
%- our method:  scalable... takes advantage of randomized algorithms and Nystrom
%- flexibility:   rapid evaluation of tuning parameter space
%- further advantages:  particular amenable to dyn. systems and long time series data.
%- we show -  Lorenz....other systems 
%
\end{abstract}

\section{Motivation}

%-
%
%- diffusion emerging:   flexible, data-driven and nonlinear manifolds
%
%- physically inspired by stochastic dyn. sys.
%
%-  current limitations:  not scalable to big data, tuning required of parameter space
%
%- our method:  scalable... takes advantage of randomized algorithms and Nystrom
%
%- Importnantly, Nystrom is exploited in a variety of kernel techniques (SVM...), but not in diff. maps
%
%- flexibility:   rapid evaluation of tuning parameter space
%
%- further advantages:  particular amenable to dyn. systems and long time series data.

In the era of `big data', dimension reduction is critical for data science. 
The aim is to map a set of high-dimensional points $x_1,x_2,...,x_n \in \mathcal{X}$ to a lower dimensional (feature) space $\mathcal{F}$
\begin{equation*}
\Psi: \mathcal{X} \subseteq \mathbb{R}^{p} \rightarrow  \mathcal{F} \subseteq \mathbb{R}^{d}, \qquad d \ll p.
\end{equation*}
The map $\Psi$ aims to preserve large scale features, while suppressing uninformative variance (fine scale features) in the data~\cite{MAL-002,van2009dimensionality}. 
%
%The hope is, that the low-dimensional points in $\mathcal{F}$ help to understand the data better and turn out to be useful features for subsequent tasks such as clustering and classification.
%
Diffusion maps provide a flexible and data-driven framework for non-linear dimensionality reduction~\cite{lafon2004diffusion,Coifman7426,Coifman2006acha,Coifman2008mmas,Nadler2006acha}. 
%
%
%The approach is in sharp contrast with dimension reduction techniques such as principal component analysis. Diffusion maps learn \LMcomment{localized features at scale} and allow to reorganize data based on the underlying geometry of the input data.
%
Inspired by stochastic dynamical systems, diffusion maps have been used in a diverse set of applications including 
face recognition~\cite{barkan2013fast}, image segmentation~\cite{Karacan:2013:SIS:2508363.2508403}, gene expression analysis~\cite{xu2007applications}, and anomaly detection~\cite{mishne2013multiscale}. 
Because computing the diffusion map scales with the number of observations $n$, it is computationally intractable for long time series data, especially as parameter tuning is also required.  
%
%We propose an alternative strategy for evaluating the diffusion map that exploits the Nystr\"om method.  
%The computation speed is greatly improved and rapid parameter tuning can be performed, making this an ideal method for evaluating long time series data.
%However, the computational complexity of diffusion maps scales with the number of observations $n$. Thus, long time series (\textit{e.g.}, data arising in dynamical systems) pose a computational challenge.
%
Randomized methods have recently emerged as a powerful strategy for handling `big data'~\cite{halko2011rand,Mahoney2011,liberty2013simple,erichson2016randomized,erichson2017compressed} and for linear dimensionality reduction~\cite{halko2011algorithm,rokhlin2009randomized, ERICHSON20181,erichson2017randomized,erichson2017randomizedCP}, with  the Nystr\"om method being a popular randomized technique for the fast approximation of kernel machines~\cite{NIPS2000_1866,drineas2005nystrom}. 
Specifically, the Nystr\"om method takes advantage of low-rank structure and a rapidly decaying eigenvalue spectrum of symmetric kernel matrices.  Thus the memory and computational burdens of kernel methods can be substantially eased.
Inspired by these ideas, we take advantage of randomization as a computational strategy and propose a Nystr\"om-accelerated diffusion map algorithm. 
%
%We show that our proposed algorithms allow the fast computation of the dominant diffusion map components.
%
%We achieve a speedup of roughly two to four times when approximating the dominant diffusion map components.  

\section{Diffusion Maps in a nutshell}

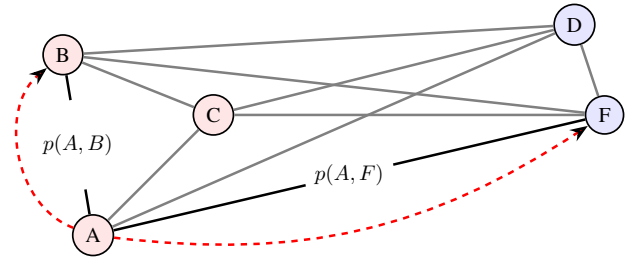
\begin{figure}[!b]
	\begin{center}\scalebox{0.80}{
			%\vspace{0.1in}
			\begin{tikzpicture}
			\begin{scope}[every node/.style={circle,thick,draw}]
			\node[fill=red!10] (A) at (-1.5, 0) {A};
			\node[fill=red!10] (B) at (-2.0, 3) {B};
			\node[fill=red!10] (C) at ( 0.5, 2) {C};
			\node[fill=blue!10] (D) at (6.5,3.5) {D};
			\node[fill=blue!10] (F) at (7,2) {F} ;
			\end{scope}
			
			\begin{scope}[>={Stealth[black]},
			every node/.style={fill = white,circle}]
			% every edge/.style={draw = black,very thick}]
			\path [-, very thick, black] (A) edge node {$p(A,B)$} (B);
			\path [-, very thick, gray] (B) edge  (C);
			\path [-, very thick, black] (A) edge node {$p(A,F)$} (F);
			\path [-, very thick, gray] (D) edge (C);
			\path [-, very thick, gray] (D) edge (F);
			\path [-, very thick, gray] (D) edge (B);
			\path [-, very thick, gray] (D) edge (A);
			
			\path [-, very thick, gray] (A) edge  (C);
			\path [-, very thick, gray] (B) edge  (F);
			
			% \path [-, very thick] (D) edge node {$a(D,F)$} (F);
			\path [-, very thick, gray] (C) edge  (F);
			
			\path [->, red, very thick, dashed] (A) edge[bend right=-60 red]  (B); 
			\path [->, red, very thick, dashed] (A) edge[bend right=20 red]  (F); 
			
			\end{scope}
			\end{tikzpicture}}
	\end{center}
	\caption{ %Diffusion maps can be related to random walks on graphs. 
		Nodes which have a high transition probability are considered to be highly connected. For instance, it is more likely to jump from node $A$ to $B$ than from $A$ to $F$. %Hence, clusters are identified as regions which are highly connected and which have a low probability of escape.
	}
	\label{Fig:random_walk}
\end{figure}

Diffusion maps explore the relationship between heat diffusion and random walks on undirected graphs. 
A graph can be constructed from the data using a kernel function $\kappa(x,y): \mathcal{X} \times \mathcal{X} \rightarrow  \mathbb{R}$, which measures the similarity for all points in the input space $x,y \in {\mathcal{X}}$.
%
% \begin{equation*}
%  \kappa: \mathcal{X} \times \mathcal{X} \rightarrow  \mathbb{R}.
% \end{equation*}
%
A similarity measure is, in some sense, the inverse of a distance function, \textit{i.e.}, similar objects take large values.
Therefore, different kernel functions capture distinct features of the data.
Given such a graph, the connectivity between two data points can be quantified in terms of the probability $p(x,y)$ of jumping from $x$ to $y$. This is illustrated in Fig.~\ref{Fig:random_walk}. 
%
%In other words, it is more likely that we jump to a nearby data-point than to a data point that is far away. This is. 
%
Specifically, the quantity $p(x,y)$ is defined as the normalized kernel 
\begin{equation}
p(x,y) := \dfrac{\kappa(x,y)}{\nu(x)}.
\end{equation}
This is known as normalized graph Laplacian construction~\cite{chung1997spectral}, where $\nu(x)$ is defined as a measure ${\nu(x)} = \int_{\mathcal{X}}  \kappa(x,y) \, \mu(y) \, \mathrm{d} y$ of degree in a graph so that we have
%
%\begin{equation}
%\int\limits_\mathcal{X} \! p(x,y) \,  \mathrm{d} \mu(y) = 1,
%\end{equation}
%
%
\begin{equation}
\int\limits_\mathcal{X} \! p(x,y) \, \mu (y) \, \mathrm{d} y = 1,
\end{equation}
where $\mu(\cdot)$ denotes the measure of distribution of the data points on $\mathcal{X}$.
This means that $p(x,y)$ represents the transition kernel of a reversible Markov chain on the graph, \textit{i.e.}, $p(x,y)$ represents the one-step transition probability from $x$ to $y$.
%
%
%In other words, the probability that a random walk which starts at $x$ reaches $y$ after $t$ steps is $p^{t}(x,y)$.
%
Now, a diffusion operator $\mathbf{P}$ can be defined by integrating over all paths through the graph as
\begin{equation}\label{eq:P}
\mathbf{P}f(x) := \int\limits_{\mathcal{X}} \! p(x,y) \, f(y) \, \mu(y) \, \mathrm{d} y, \qquad \: \forall f \, \in \, L_1\left(\mathcal{X}\right),
\end{equation}
so that $\mathbf{P}$ defines the entire Markov chain~\cite{Coifman7426}.
More generally, we can define the probability of transition from each point to
another by running the Markov chain $t$ times forward:
\begin{equation}
\mathbf{P}^{t}f(x) := \int\limits_X \! p^{t}(x,y) \, f(y) \, \mu(y) \, \mathrm{d} y.
\end{equation}
The rationale is that the underlying geometric structure of the dataset is revealed at a magnified scale by taking larger powers of $\mathbf{P}$. 
Hence, the diffusion time $t$ acts as a scale, \textit{i.e.}, the transition probability between far away points is decreased with each time step $t$. 
%
% Recall, the probability that a random walk which starts at $x$ reaches $y$ after $t$ steps is $p^{t}(x,y)$. Now, this probability is provided by the diffusion operator $\mathbf{P}^{t}$. 
%
Spectral methods can be used to characterize the properties of the Markov chain.
To do so, however, we need to define first a symmetric operator $\mathbf{A}$ as
\begin{equation}\label{eq:A}
\mathbf{A}f(x) := \int\limits_\mathcal{X}  \! {a}(x,y) \, f(y) \, \mu(y) \, \mathrm{d} y
\end{equation}
by normalizing the kernel with a symmetrized measure
\begin{equation}
{a}(x,y) := \dfrac{\kappa(x,y)}{ \sqrt{\nu(x)} \sqrt{\nu(y)}}.
\end{equation}
This ensures that ${a}(x,y)$ is symmetric, ${a}(x,y)={a}(y,x)$, and positivity preserving $a(x,y)\geq 0\,\,\forall x,y$~\cite{Coifman2006acha,lovasz1993random}.
Now, the eigenvalues $\lambda_i$ and corresponding eigenfunctions $\phi_i(x)$ of the operator $\mathbf{A}$ can be used to describe the transition probability of the diffusion process.
Specifically, we can define the components of the diffusion map $\Psi^t(x)$ as the scaled eigenfunctions of the diffusion operator
%
%\begin{equation}
%\Psi^t(x) = \left( \begin{array}{c}
%\sqrt{\lambda_1^{t}} \phi_1(x)\\ 
%\sqrt{\lambda_2^{t}} \phi_2(x)\\ 
%\vdots \\ 
%\sqrt{\lambda_n^{t}} \phi_n(x) 
%\end{array} \right).
%\end{equation}
% 
\begin{equation*}
\Psi^t(x) = \left(\sqrt{\lambda_1^{t}} \, \phi_1(x), \sqrt{\lambda_2^{t}} \, \phi_2(x),...,\sqrt{\lambda_n^{t}} \, \phi_n(x) \right).
\end{equation*}
The diffusion map $\Psi^t(x)$ captures the underlying geometry of the input data.
Finally, to embed the data into an Euclidean space, we can use the diffusion map to evaluate the diffusion distance between two data points
\begin{equation*}
D^2_t(x,y)=||\Psi^t(x) - \Psi^t(y)||^2 \approx \sum_{i=1}^{d} \lambda_i^t (\phi_i(x) - \phi_i(y))^2,
\end{equation*}
where we may retain only the $d$ dominant components to achieve dimensionality reduction. 
The diffusion distance reflects the connectivity of the data, \textit{i.e.}, points which are characterized by a high transition probability are considered to be highly connected.
This notion allows one to identify clusters in regions which are highly connected and which have a low probability of escape~\cite{lafon2004diffusion,Coifman2006acha}. 
%
%
%to the 
%
%Data can be mapped into an Euclidean space, where the Euclidean distance between two mapped points $Y_{i}$ and $Y_{j}$ is the diffusion distance between data points $X_{i}$ and $X_{j}$ in data space. 
%

\section{Diffusion Maps meet Nystr\"om}

The Nystr\"om method~\cite{nystrom1930praktische} provides a powerful framework to solve Fredholm integral equations which take the form
\begin{equation}\label{eq:FredholmInt}
\int  \! {a}(x,y) \, \phi_i(y) \, \mu (y) \, \mathrm{d} y = \lambda_i \phi_i(x).
\end{equation}
We recognize the resemblance with~\eqref{eq:A}. Suppose, we are given a set of independent and identically distributed samples  $\{x_1, x_j, ..., x_l\}$ drawn from $\mu (y)$. Then, the idea is it to approximate Equation~\eqref{eq:FredholmInt} by computing the empirical average
\begin{equation}
\dfrac{1}{l}\sum_{j=1}^{l} {a}(x,x_j) \phi_i(x_j) \approx \lambda_i \phi_i(x).
\end{equation}
Drawing on these ideas, Williams and Seeger~\cite{NIPS2000_1866} proposed the Nystr\"om  method for the fast approximation of kernel matrices. This has led to a large body of research and we refer to~\cite{drineas2005nystrom} for an excellent and comprehensive treatment.

\subsection{Nystr\"om Accelerated Diffusion Maps Algorithm}
Let us express the diffusion maps algorithm in matrix notation. Let $\mathbf{X} \in \mathbb{R}^{n\times p}$ be a dataset with $n$ observations and $p$ variables. 
Then, given $\kappa$ we form a symmetric kernel matrix $\mathbf{K} \in \mathbb{R}^{n\times n}$ where each entry is obtained as $\mathbf{K}_{{i,j}} = \kappa(x_{i},x_{j})$.

The diffusion operator in Equation~\eqref{eq:P} can be expressed in the form of a diffusion matrix $\mathbf{P} \in \mathbb{R}^{n\times n}$ as 
\begin{equation}
\mathbf{P} := \mathbf{D}^{-1} \mathbf{K},
\end{equation}
where $\mathbf{D} \in \mathbb{R}^{n\times n}$ is a diagonal matrix which is computed as $\mathbf{D}_{{i,i}} = \sum _{j} \mathbf{K}_{{i,j}}$.
Next, we form a symmetric matrix
\begin{equation}
\mathbf{A} := \mathbf{D}^{-\frac{1}{2}} \mathbf{K} \mathbf{D}^{-\frac{1}{2}},
\end{equation}
which allows us to compute the eigendecomposition 
\begin{equation}
\mathbf{A} = \mathbf{U} \mathbf{\Lambda} \mathbf{U}^\top.
\end{equation}
The columns $\phi_i \in \mathbb{R}^{n}$ of $\mathbf{U} \in \mathbb{R}^{n\times n}$ are the orthonormal eigenvectors. 
The diagonal matrix $\mathbf{\Lambda}\in \mathbb{R}^{n\times n}$ has the eigenvalues $\lambda_1 \geq \lambda_2 \geq ... \geq 0$ in descending order as its entries. 
%
%The largest eigenvalue of both $\mathbf{P}$ and $\mathbf{A}$ is $\lambda_1=1$~\cite{lovasz1993random}.

The Nystr\"om method can now be used to quickly produce an approximation for the dominant $d$ eigenvalues and eigenvectors~\cite{NIPS2000_1866}. Assuming that $\mathbf{A} \in \mathbb{R}^{n\times n}$ is a symmetric positive semidefinite matrix (SPSD), the Nystr\"om method yields the following low-rank approximation for the diffusion matrix
\begin{equation}\label{eq:mat_nystroem}
\mathbf{A} \approx \mathbf{C} \mathbf{W}^{-1} \mathbf{C}^\top,
\end{equation}
where $\mathbf{C}$ is an $n\times d$ matrix which approximately captures the row and column space of  $\mathbf{A}$. The matrix $\mathbf{W}$ has dimension $d\times d$ and is SPSD.
%
%Strategies to obtain the matrices $\mathbf{C}$ and $\mathbf{W}$ are discussed below.
%
Following, Halko et al. \cite{halko2011rand}, we can factor $A$ in Equation~\eqref{eq:mat_nystroem} using the Cholesky decomposition
\begin{equation}
\mathbf{A} \approx \mathbf{F}\mathbf{F}^\top,
\end{equation}
where $\mathbf{F} \in \mathbb{R}^{n\times d}$ is the approximate Cholesky factor, defined as
$\mathbf{F} := \mathbf{C} \mathbf{W}^{-\frac{1}{2}}$.
Then, we can obtain the eigenvectors and eigenvalues by computing the singular value decomposition
\begin{equation}
\mathbf{F} = \mathbf{\tilde{U}} \mathbf{\Sigma} \mathbf{V}^\top.
\end{equation}
The left singular vectors $\mathbf{\tilde{U}} \in \mathbb{R}^{n\times d}$ are the dominant $d$ eigenvectors of $\mathbf{A}$ and $\mathbf{\Lambda}=\mathbf{\Sigma}^2 \in \mathbb{R}^{d\times d}$ are the corresponding $d$ eigenvalues. Finally, we can recover the eigenvectors of the diffusion matrix $\mathbf{P}$ as $\mathbf{U = D\tilde{U}}$.

\subsection{Matrix Sketching}

Different strategies are available to form the matrices $\mathbf{C}$ and $\mathbf{W}$. Column sampling is most computational and memory efficient. Random projections have the advantage that they often provide an improved approximation. Thus, the different strategies pose a trade-off between speed and accuracy and the optimal choice depends on the specific application.

\subsubsection{Column Sampling}

The most popular strategy to form $\mathbf{C} \in \mathbb{R}^{n\times d}$ is column sampling, \textit{i.e.}, we sample $d$ columns from $\mathbf{A}$. 
Subsequently, the small matrix $\mathbf{W} \in \mathbb{R}^{d\times d}$ is formed by extracting $d$ rows from $\mathbf{C}$. Given an index vector $J\in \mathbb{N}^{d}$ we form the matrices as
\begin{equation}
\mathbf{C} := \mathbf{A}(:,J) \quad \text{and} \quad \mathbf{W} := \mathbf{C}(J,:) = \mathbf{A}(J,J).
\end{equation}
The index vector can be designed using random (uniform) sampling or importance sampling~\cite{kumar2012sampling}.
Column sampling is most efficient, because it avoids explicit construction of the kernel matrix. 
%
%It follows that $\mathbf{W}$ is SPSD, since it is formed of the entries of $\mathbf{A}$ where the $J$ columns and rows are intersecting.
%
For details we refer to~\cite{drineas2005nystrom}. 

\subsubsection{Random Projections}

The second strategy is to use random projections\cite{halko2011rand}. First, we form a random test matrix $\mathbf{\Omega} \in \mathbb{R}^{n\times l}$ which is used to sketch the diffusion matrix
\begin{equation}\label{eq:sketch}
\mathbf{S} := \mathbf{A} \mathbf{\Omega}.
\end{equation}
where $l\ge d$ is slightly larger than the desired target rank $d$. 
Due to symmetry, the columns of $\mathbf{S} \in \mathbb{R}^{n\times l}$ provide a basis for both the column and row space of $\mathbf{A}$. 
Then, an orthonormal basis  $\mathbf{Q} \in \mathbb{R}^{n\times l}$ is obtained by computing the QR decomposition as $\mathbf{S} = \mathbf{Q} \mathbf{R}$.
We form the matrix $\mathbf{C} \in \mathbb{R}^{n\times l}$ and $\mathbf{W} \in \mathbb{R}^{l\times l}$ by projecting $\mathbf{A}$ to a lower-dimensional space as
\begin{equation}
\mathbf{C} := \mathbf{A} \mathbf{Q} \quad \text{and} \quad \mathbf{W} := \mathbf{Q}^\top \mathbf{C}.
\end{equation}
% 

%This approach requires two passes over the input matrix. 

Further, the power iteration scheme can be used to improve the quality of the basis matrix $\mathbf{Q}$~\cite{halko2011rand}.
The idea is to sample from a preprocessed matrix
$\mathbf{S} = (\mathbf{A}\mathbf{A}^\top)^q \mathbf{A} \mathbf{\Omega}$,
instead of directly sampling from $\mathbf{A}$ as in Equation~\eqref{eq:sketch}. 
Here, $q$ denotes the number of power iterations. In practice, this is implemented efficiently via subspace iterations.

%
%we define the normalized kernel as
%%
%\begin{equation}
%\mathbf{A}_{{i,j}} = \frac {\mathbf{K}_{{i,j}}}{\nu(x_{i})^{\alpha } \nu(x_{j})^{\alpha }  }.
%\end{equation}
%%
%This is equivalently to
%%
%\begin{equation}
%\mathbf{A} = \mathbf{D}^{{-\alpha }} \mathbf{K} \mathbf{D}^{{-\alpha }},
%\end{equation}
%%
%where $\mathbf{D}$ is a diagonal matrix and its entries are the column sums of the normalized kernel $D_{{i,i}}=\sum _{j} \mathbf{A}_{{i,j}}$.
%
%The reason to introduce the normalization step involving $\alpha$ is to tune the influence of the data point density on the infinitesimal transition of the diffusion. In some applications, the sampling of the data is generally not related to the geometry of the manifold we are interested in describing. In this case, we can set $\alpha =1$ and the diffusion operator approximates the Laplace–Beltrami operator. We then recover the Riemannian geometry of the data set regardless of the distribution of the points. To describe the long-term behavior of the point distribution of a system of stochastic differential equations, we can use α = 0.5 $\alpha = 0.5$ and the resulting Markov chain approximates the Fokker-Planck diffusion. With $\alpha = 0$, it reduces to the classical graph Laplacian normalization.

\section{Results}

In the following, we demonstrate the efficiency of our proposed Nystr\"om accelerated diffusion map algorithm. First, we explore both toy data and time-series data from a dynamical system. Then, we evaluate the computational performance and compare it with the deterministic diffusion map algorithm. %\footnote{The deterministic algorithm uses the SciPy eigensolver for sparse matrices.}
Here, we restrict the evaluation to the Gaussian kernel:
\begin{equation*}
\kappa(x,y) = \exp\left( -\sigma^{-1} ||x-y||_2^2\right),
\end{equation*}
where $\sigma$ controls the variance (width) of the distribution. 
%

%In addition, delay coordinates have also been used to discover embeddings of nonlinear systems in which the dynamics become approximately linear~\cite{Brunton2017natcomm,Das2017arxiv,Arbabi2016arxiv}. 

\subsection{Artificial Toy Datasets}

\begin{figure}[!h]
	\centering
	\begin{subfigure}[t]{0.20\textwidth}
		\centering
		\DeclareGraphicsExtensions{.pdf}
		\includegraphics[width=1\textwidth]{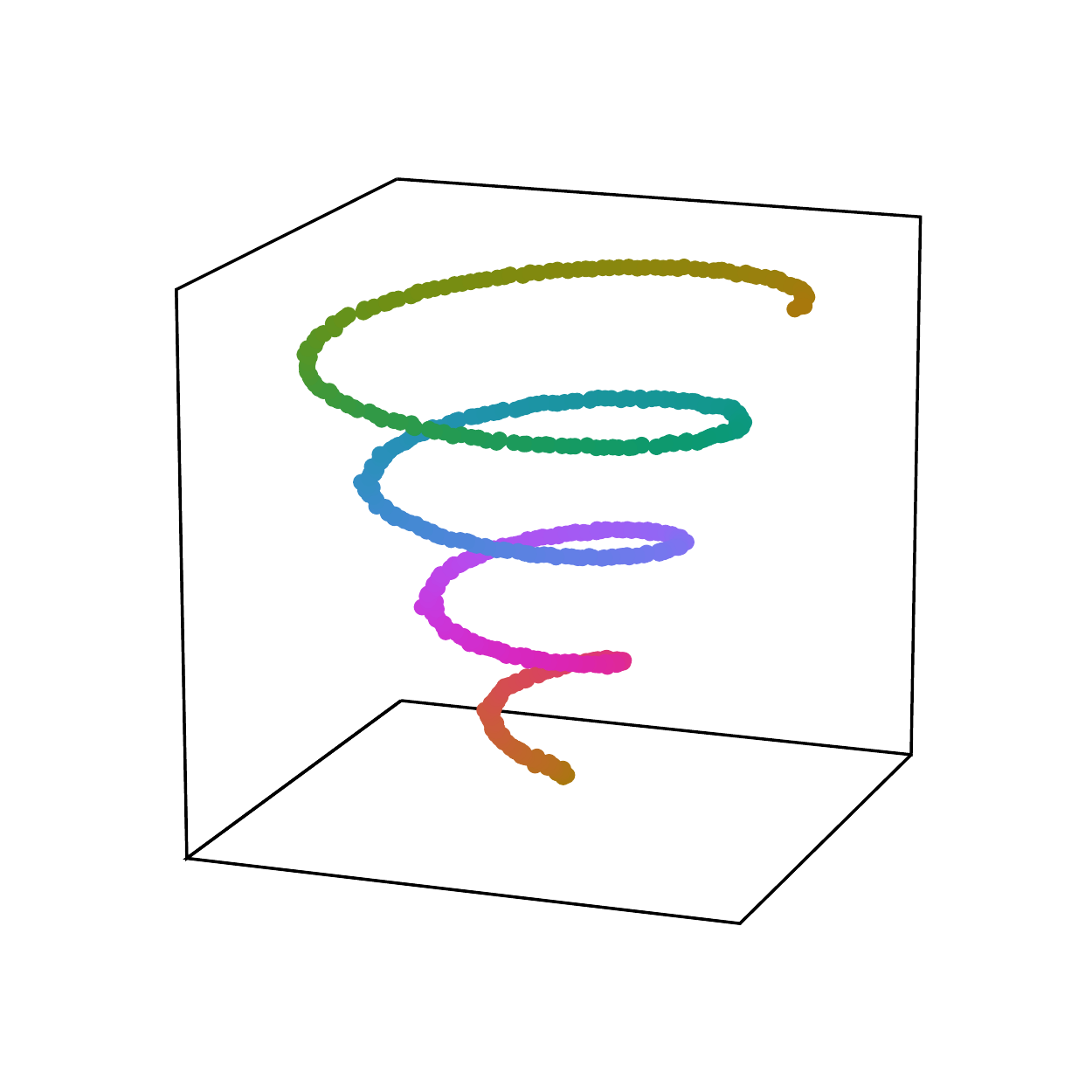}
		%\caption{Nosiy Helix}
	\end{subfigure}
	~
	\begin{subfigure}[t]{0.20\textwidth}
		\centering
		\DeclareGraphicsExtensions{.pdf}
		\includegraphics[width=1\textwidth]{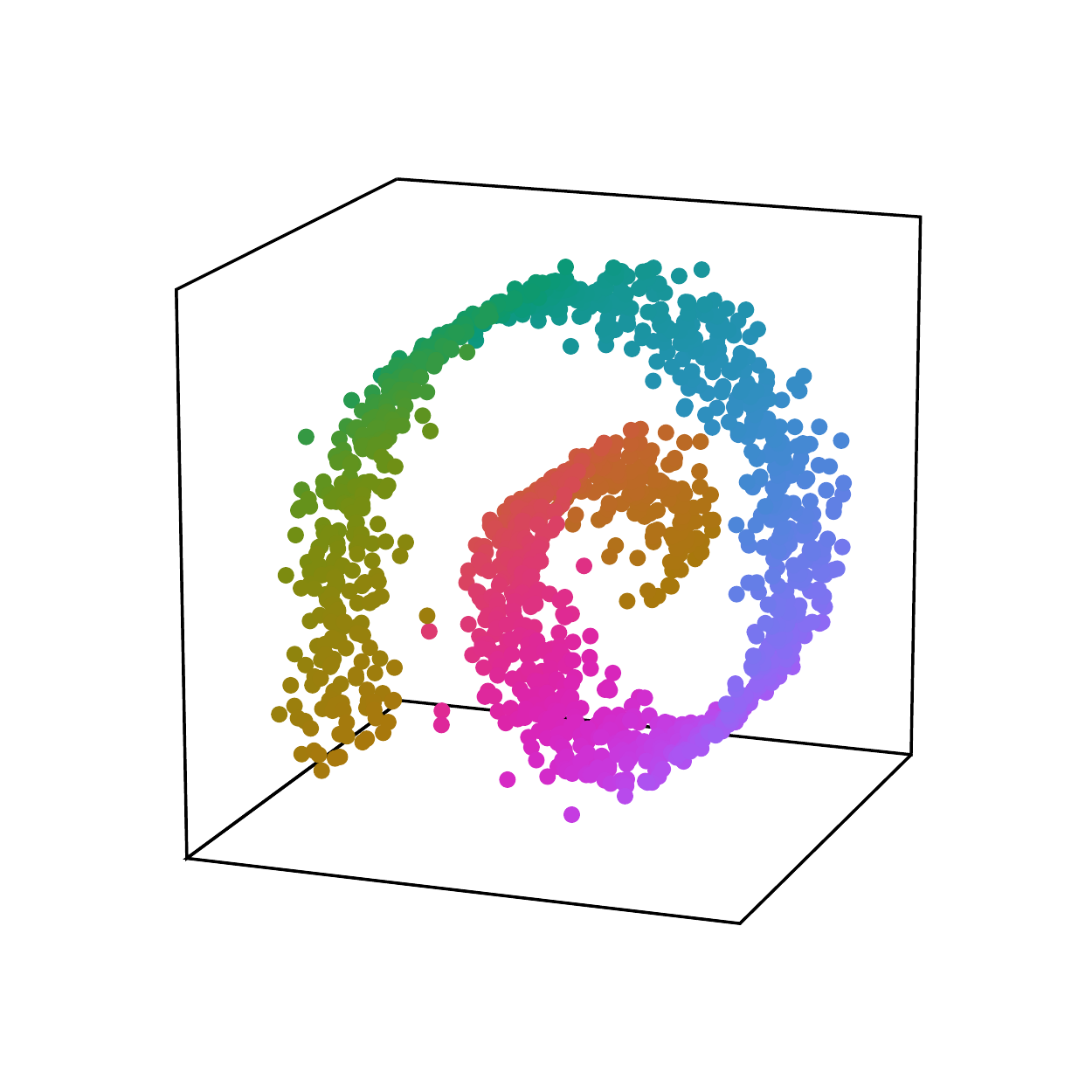}
		%\caption{Noisy Swiss-role}
	\end{subfigure}

	\begin{subfigure}[t]{0.18\textwidth}
		\centering
		\DeclareGraphicsExtensions{.pdf}
		\includegraphics[width=1\textwidth]{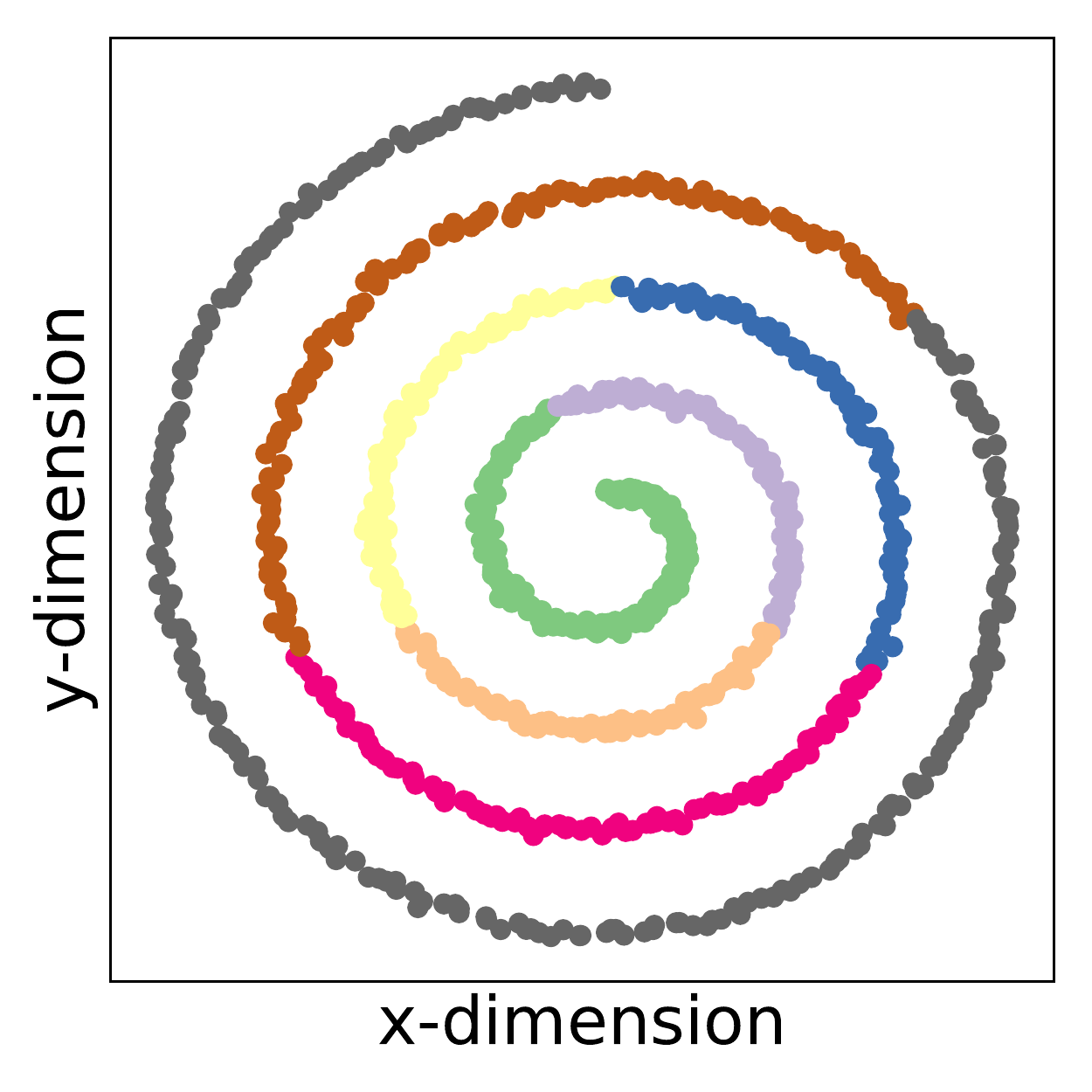}
	\end{subfigure}
	~
	\begin{subfigure}[t]{0.18\textwidth}
		\centering
		\DeclareGraphicsExtensions{.pdf}
		\includegraphics[width=1\textwidth]{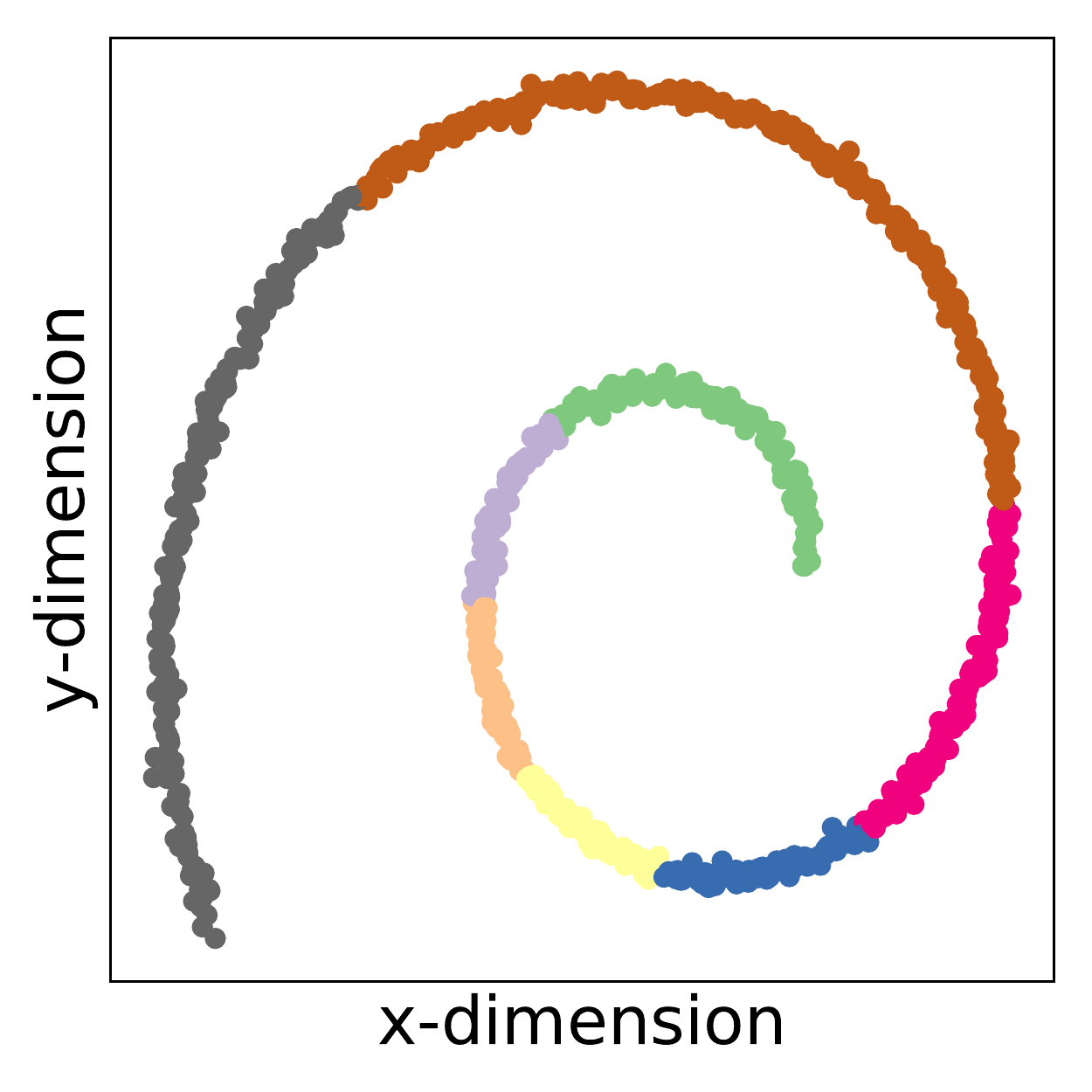}

	\end{subfigure}	
	
	\begin{subfigure}[t]{0.18\textwidth}
		\centering
		\DeclareGraphicsExtensions{.pdf}
		\includegraphics[width=1\textwidth]{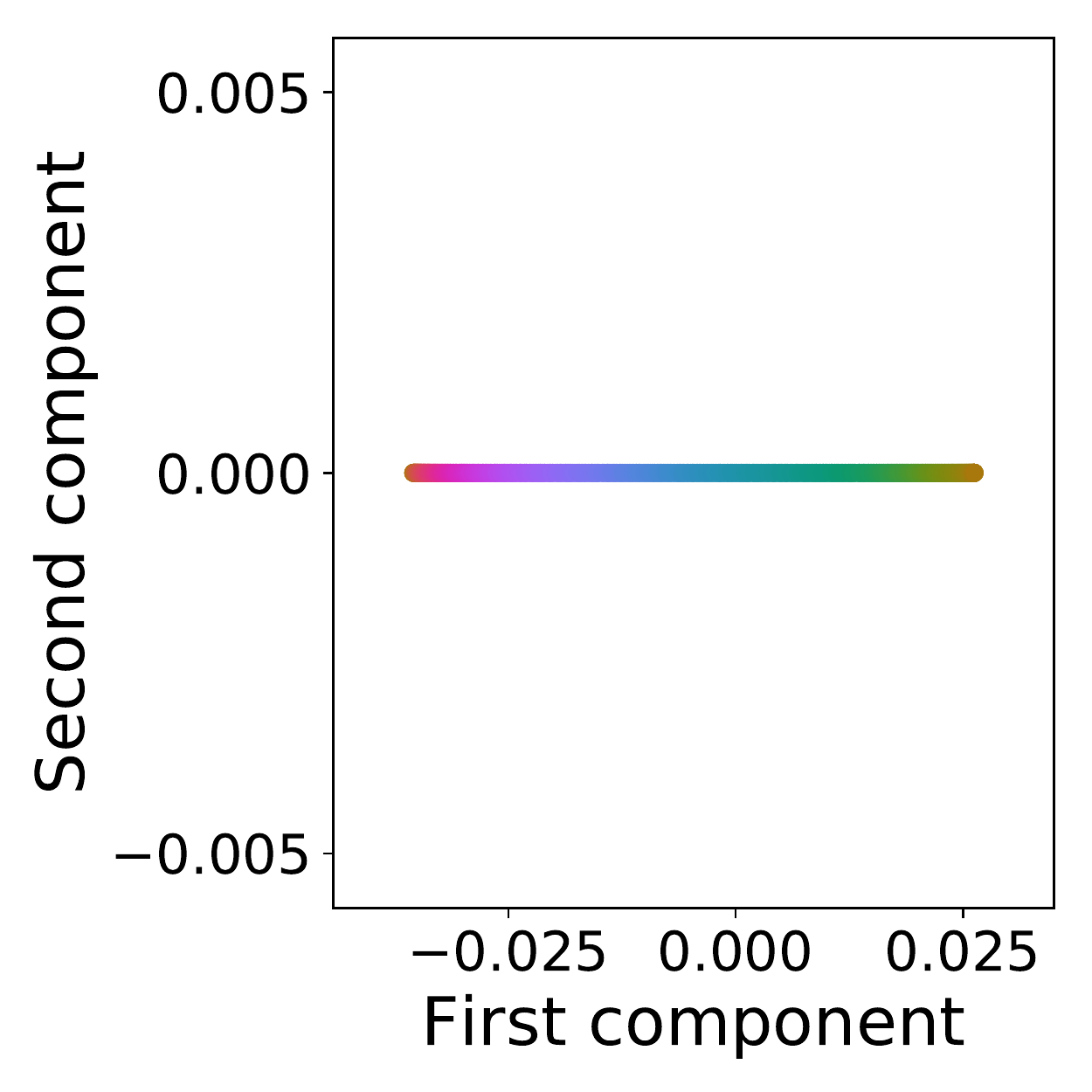}
		\caption{Noisy helix}
	\end{subfigure}			
	~
	\begin{subfigure}[t]{0.18\textwidth}
		\centering
		\DeclareGraphicsExtensions{.pdf}
		\includegraphics[width=1\textwidth]{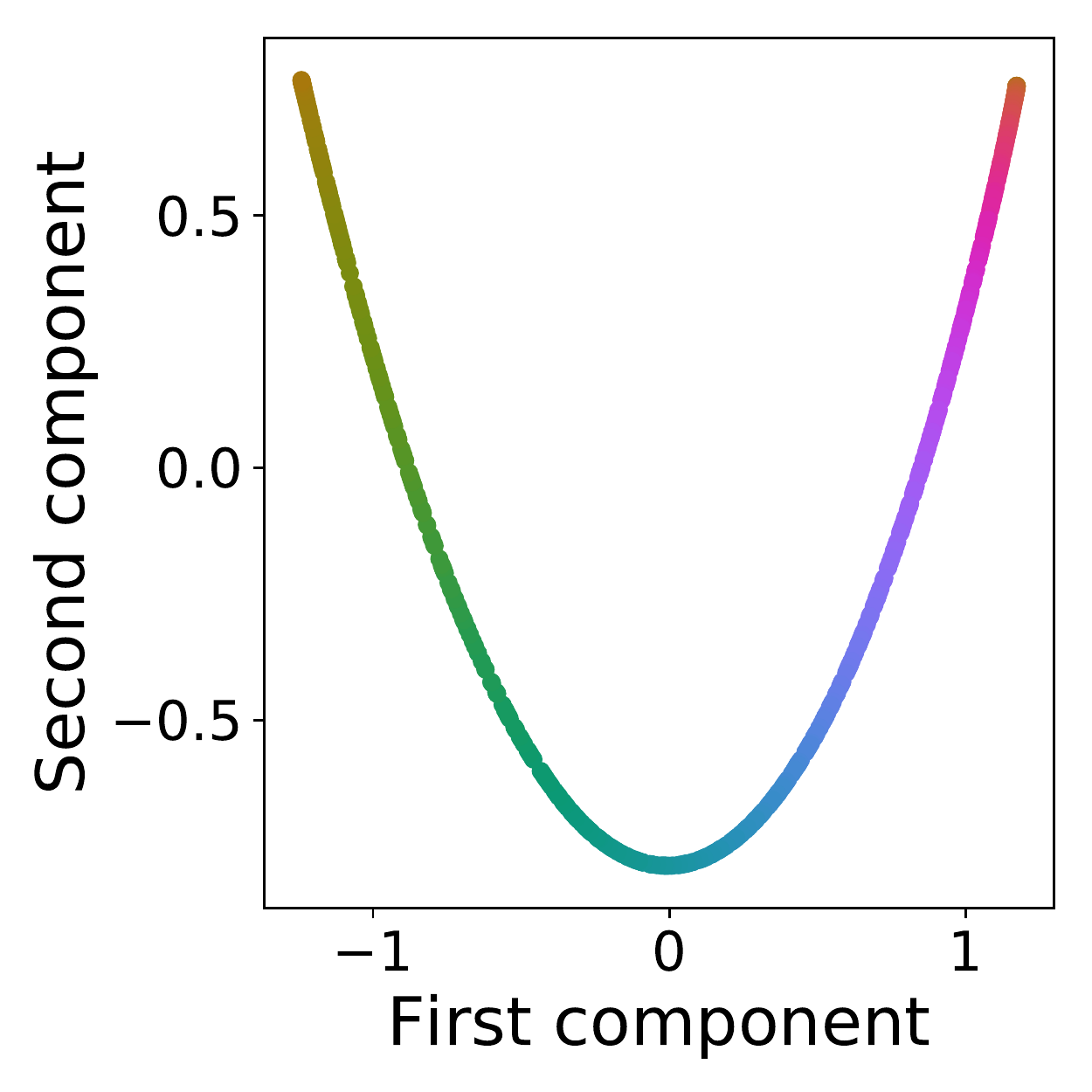}
		\caption{Noisy Swiss role}
	\end{subfigure}			
	
	\caption{The top row shows the datasets. The second row shows the clustered data points at diffusion time $t=100$. The third row shows low-dimensional embedding computed using the Nystr\"om-accelerated diffusion map algorithm.}
	\label{fig:toy_data}
\end{figure}

First, we consider two non-linear artificial datasets: the helix and the famous Swiss role dataset. 
Both datasets are perturbed with a small amount of white Gaussian noise. 
Figure~\ref{fig:toy_data} shows both datasets. The first two components of the diffusion map $\Psi^t(x)$ are used to illustrate the non-linear embedding in two dimensional space at time $t=100$. Then, we use the diffusion distance to cluster the data points. Indeed, the diffusion map is able to correctly cluster both non-linear data sets. The width of the Gaussian kernel is set to $\sigma=0.5$. 

\newpage
\subsection{The Chaotic Lorenz System}
 
\begin{figure}[!bt]
	\centering
	\begin{subfigure}[t]{0.2\textwidth}
		\centering
		\DeclareGraphicsExtensions{.pdf}
		\includegraphics[width=1\textwidth]{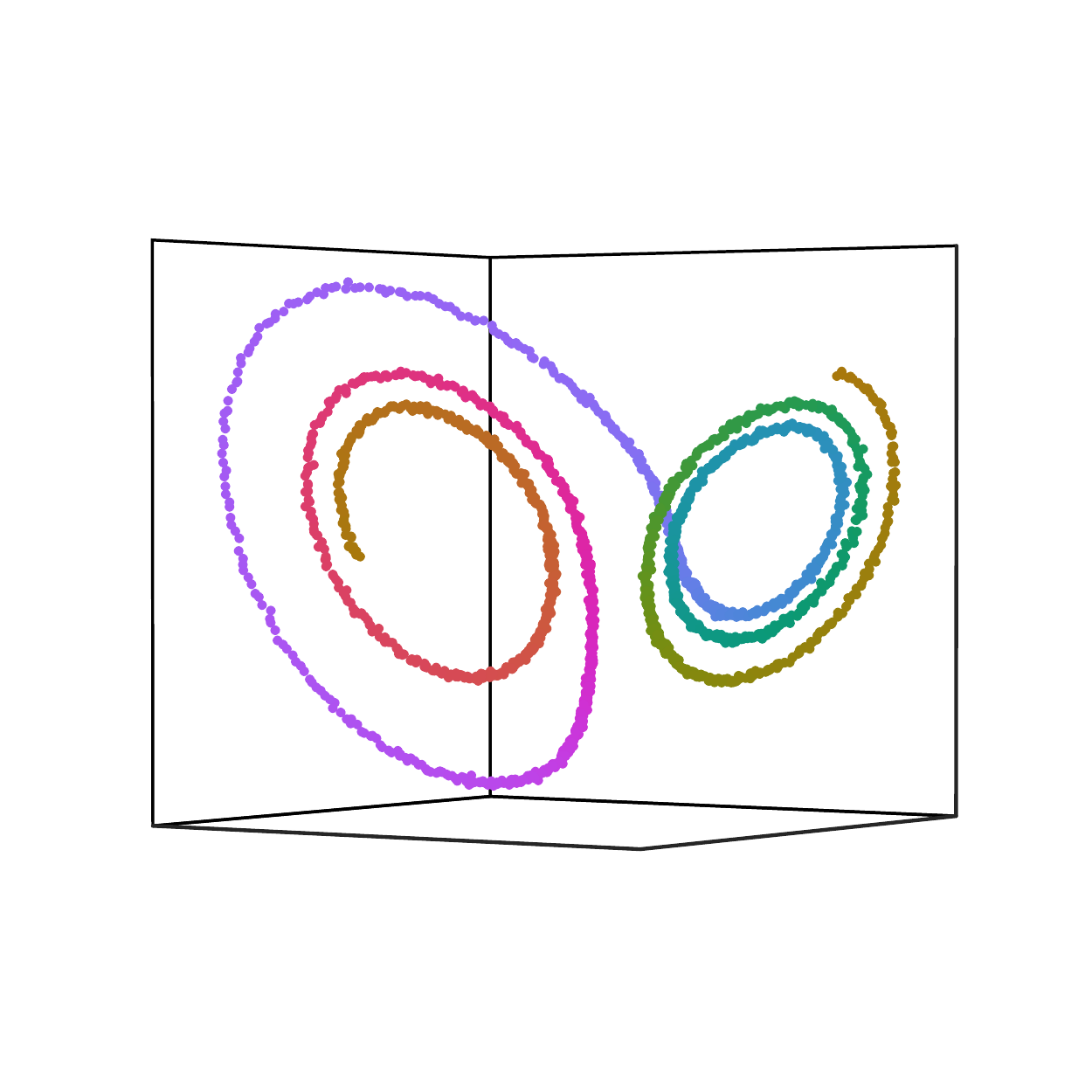}
		\caption{Lorenz system}
	\end{subfigure}
	~
	\begin{subfigure}[t]{0.18\textwidth}
		\centering
		\DeclareGraphicsExtensions{.pdf}
		\includegraphics[width=1\textwidth]{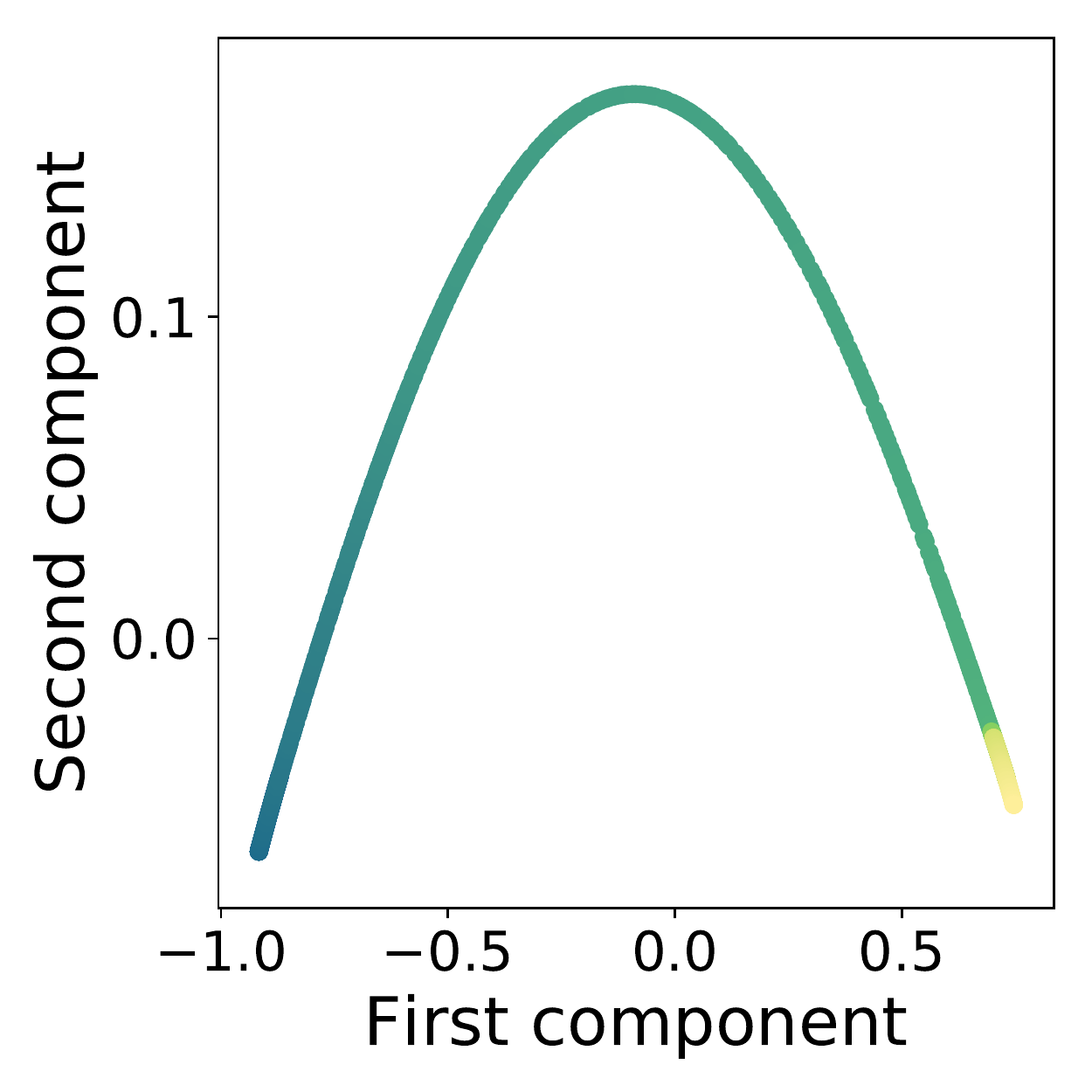}
		\caption{2-D embedding}
	\end{subfigure}			
	
	\begin{subfigure}[t]{0.18\textwidth}
		\centering
		\DeclareGraphicsExtensions{.pdf}
		\includegraphics[width=1\textwidth]{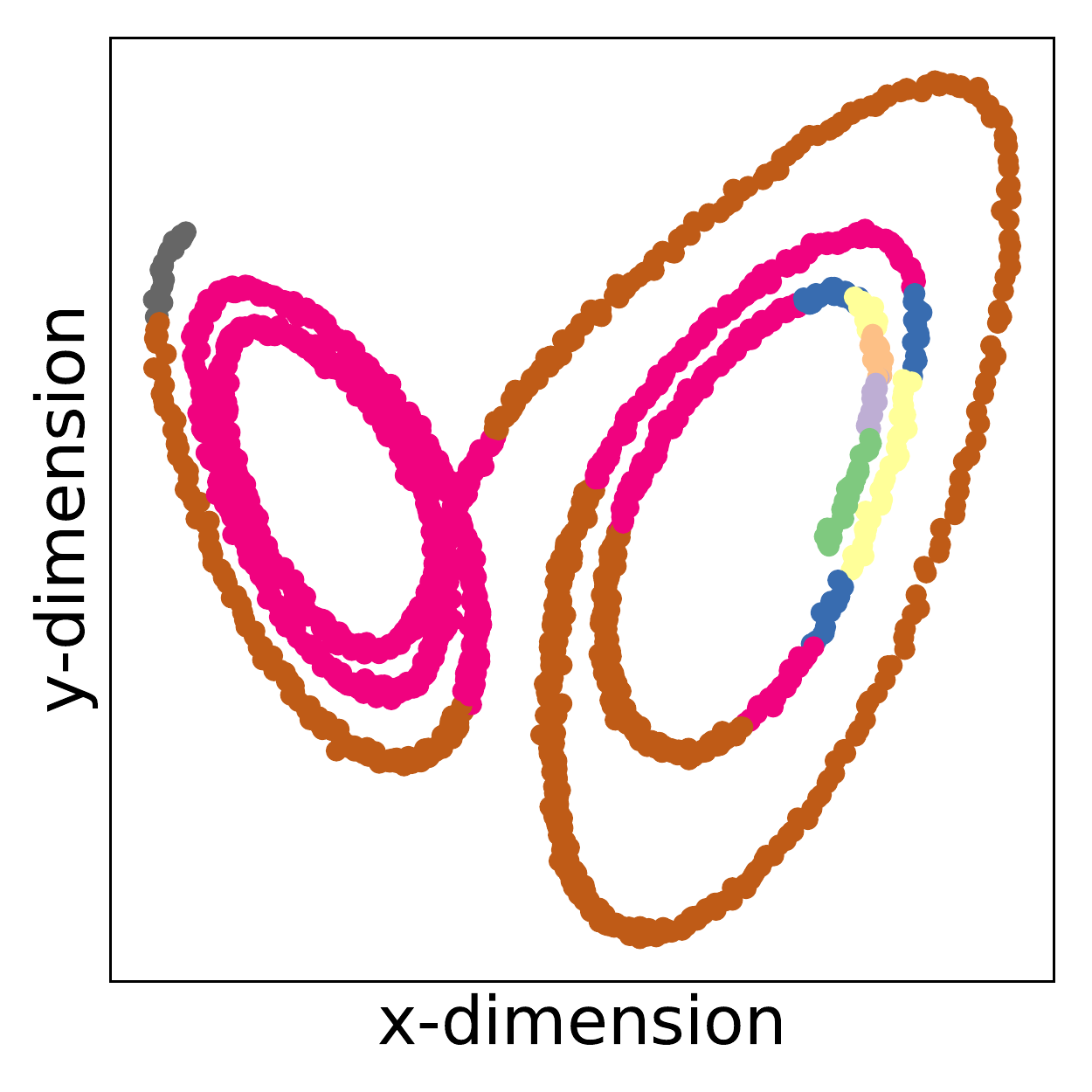}
		\caption{Clustered: $t=100$}
	\end{subfigure}		
	~
	\begin{subfigure}[t]{0.18\textwidth}
		\centering
		\DeclareGraphicsExtensions{.pdf}
		\includegraphics[width=1\textwidth]{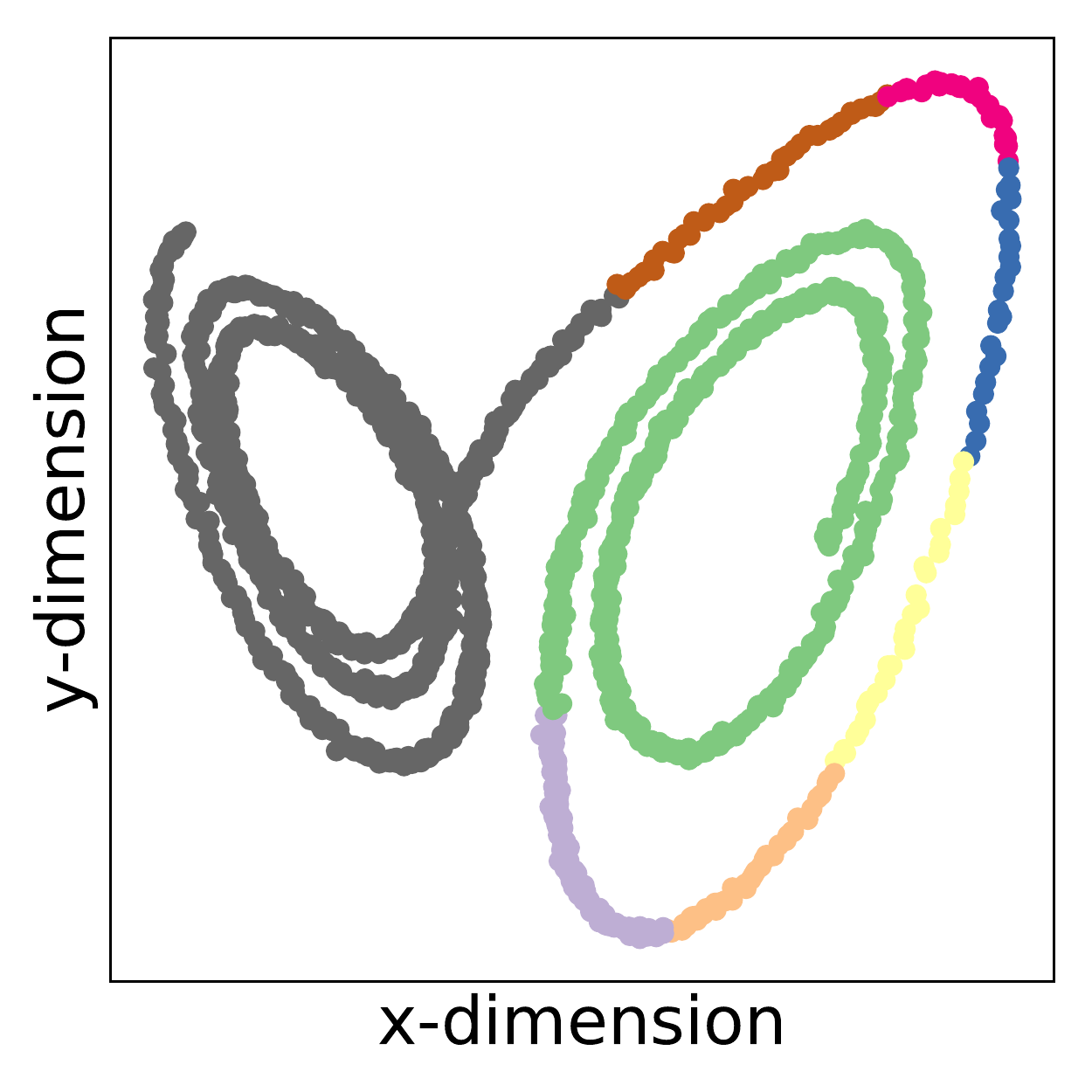}
		\caption{Clustered: $t=10000$}
	\end{subfigure}
	
	\caption{The chaotic Lorenz system and its two-dimensional embedding using diffusion maps. Here, a large number of diffusion time steps $t$ is required to obtain stable clusters.}
	\label{Fig:lorenz}
\end{figure}

Next, we explore the embedding of nonlinear time-series data.
Discovering nonlinear transformations that map dynamical systems into a new coordinate system with favorable properties is at the center of modern efforts in data-driven dynamics.  
One such favorable transformation is obtained by eigenfunctions of the Koopman operator, which provides an infinite-dimensional but linear representation of nonlinear dynamical systems~\cite{Koopman1931pnas,Mezic2005nd,Brunton2016plosone}.  
%
%Diffusion maps have recently been connected to Koopman analysis by Giannakis~\cite{Giannakis2015arxiv}, and have also been used to discover normal forms for dynamical systems simply from unlabled trajectory data~\cite{Yair2017pnas}.   
%
Diffusion maps have recently been connected to Koopman analysis and are now increasingly being employed to analyze coherent structures and nonlinear embeddings of dynamical systems~\cite{Giannakis2015siads,Berry2015pre,Yair2017pnas}.  
Here, we explore the chaotic Lorenz system, which is among the simplest and well-studied chaotic dynamical system~\cite{Lorenz1963jas}:
\begin{equation}\label{Eq:Lorenz}
[\dot{x},\,\dot{y},\,\dot{z}] = [ \sigma (y - x),\, x(\rho -z) - y,\, x y - \beta z],
\end{equation}
with parameters $\sigma\!=\!10,\rho\!=\!28,$ and $\beta\!=\!8/3$.
%
%\begin{eqnarray*}\label{Eq:Lorenz}
%\dot{x} & = & \sigma (y - x)\\
%\dot{y} & = & x(\rho -z) - y\\
%\dot{z} & = & x y - \beta z,
%\end{eqnarray*}
%%
%
%We use initial conditions $\begin{bmatrix}-8 & 8 & 27\end{bmatrix}^T$ and integrate the system from $t\!=\!0$ to $t\!=\!5$ with $\Delta t\approx0.0001$, resulting in $30,000$ time steps.   
%
We use the initial conditions $\begin{bmatrix}-8 & 8 & 27\end{bmatrix}^\top$ and integrate the system from $t\!=\!0$ to $t\!=\!5$ with $\Delta t\approx0.0001$.
Figure~\ref{Fig:lorenz} shows the results. %Note, a large number of time steps $t$ is required to find stable clusters.

\subsection{Computational Performance}

Table~\ref{table:performance_summary} gives a flavor of the computational performance of the Nystr\"om-accelerated diffusion map algorithm. We achieve substantial computational savings over the deterministic diffusion map algorithm, while attaining small errors. The relative errors between the deterministic $\Psi^t(x)$ and randomized diffusion maps $\tilde{\Psi}^t(x)$ at $t\!=\!1$ are computed in the Frobenius norm: $||\, |\Psi^t(x)| - |\tilde{\Psi}^t(x)| \, ||_F / ||\, |\Psi^t(x)| \, ||_F$. 
\begin{table}[!htb]
\caption{Computational performance for both the deterministic and the Nystr\"om accelerated diffusion map algorithm.}
\centering
\scalebox{0.8}{ 
\begin{tabular}{l c c c c c c} % centered columns (4 columns)
\hline\hline %inserts double horizontal lines
Dataset & \thead{Number of \\ Observations} & \thead{Time in s\\ Deterministic} & \thead{Time in s\\ Nystr\"om} & Speedup & Error \\ [0.5ex] % inserts table
%heading
\hline % inserts single horizontal line
Helix 			&  15,000   &  40  &  11  & 3.6  &  1.8e-13\\ 
Swiss role  	&  20,000   &  72  &  19  & 3.7  &  0.001  \\ 
Lorenz   		&  30,000 	& 351  & 115  & 3.0  &  0.06 \\
[1ex] % [1ex] adds vertical space
\hline 
\end{tabular}}
\label{table:performance_summary} 
\end{table}
\begin{figure}[!h]
	\centering
	\DeclareGraphicsExtensions{.pdf}
	\includegraphics[width=0.45\textwidth]{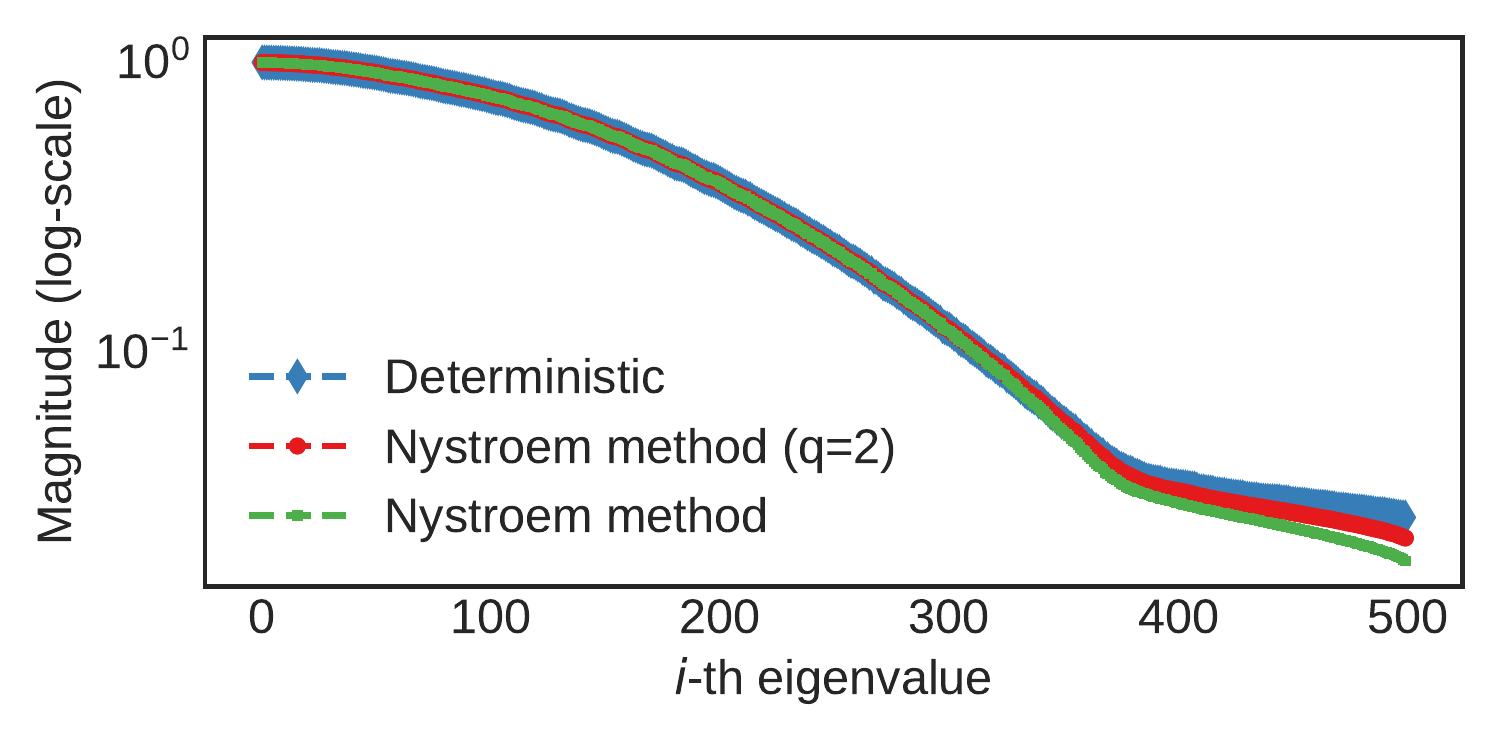}
	\caption{The Nystr\"om method faithfully captures the dominant eigenvalues of the Gaussian kernel for the Lorenz system.}
	\label{fig:spectrum}
\end{figure}

The algorithms are implemented in Python and code is available via GitHub: \url{https://github.com/erichson}. The deterministic algorithm uses the fast ARPACK eigensolver provided by SciPy. The Nystr\"om accelerated diffusion map algorithm is computed using random projections with slight oversampling and two additional power iterations $q\!=\!2$. The target-rank (number of components) is set to $d=300$ for the toy data and $d\!=\!500$ for the Lorenz system.
Figure~\ref{fig:spectrum} shows the approximated eigenvalues for the Lorenz system.

\section{Discussion}

The computational complexity of diffusion maps scales with the number of observations $n$.
Thus, applications such as the analysis of time-series data from dynamical systems pose a computational challenge for diffusion maps.
Fortunately, the Nystr\"om method can be used to ease the computational demands. 
However, diffusion maps are highly sensitive to the approximated range subspace which is provided by the eigenvectors. This means that the Nystr\"om method provides a good approximation only if: (a) the kernel matrix has low-rank structure; (b) the eigenvalue spectrum has a fast decay. 
The Nystr\"om method shows an excellent performance using random projections with additional power iterations. 
We achieve a speedup of roughly two to four times when approximating the dominant diffusion map components. 
Unfortunately, the approximation quality turns out be poor using random column sampling. 
Future research opens room for a more comprehensive evaluation study. Further, it is of interest to explore  kernel functions which are more suitable for dynamical systems, \textit{e.g.}, cone-shaped kernels~\cite{zhao1990use,Giannakis2015siads}.

% References should be produced using the bibtex program from suitable
% BiBTeX files (here: strings, refs, manuals). The IEEEbib.bst bibliography
% style file from IEEE produces unsorted bibliography list.
% -------------------------------------------------------------------------
\newpage
\bibliographystyle{IEEEbib}
\small
\begin{spacing}{0.93}
\bibliography{references}
\end{spacing}
\end{document}